\documentclass[conference,final]{IEEEtran}
\pdfoutput=1
\IEEEoverridecommandlockouts
\def\BibTeX{{\rm B\kern-.05em{\sc i\kern-.025em b}\kern-.08em
    T\kern-.1667em\lower.7ex\hbox{E}\kern-.125emX}}

\usepackage{setspace}
\usepackage[utf8]{inputenc} 
\usepackage{url}            
\usepackage{booktabs}       
\usepackage{amsfonts}       
\usepackage{microtype}      
\usepackage[T1]{fontenc}
\usepackage{subcaption}
\usepackage{graphicx}
\usepackage{relsize}
\usepackage{tikz}
\usepackage{xcolor}
\definecolor{myred}{rgb}{.7,.0,.0}
\definecolor{myblue}{rgb}{.0,.0,.7}
\usepackage{hyperref}
\usepackage[all]{hypcap}
\usepackage{amsthm,amsmath,amssymb,mathtools,sansmath}
\usepackage{latexsym,wasysym}
\usepackage{transparent}
\usepackage{fancyhdr}
\newcommand{\brc}[1]{{\relsize{-1}(#1)}}

\newcommand*\circled[1]{
  \tikz[baseline=(char.base)]{
    \node[shape=circle,draw,inner sep=.1pt] (char) {#1};
  }
}
\fancypagestyle{copyright}{%
    \fancyhf{}
    \cfoot{\footnotesize%
        978-1-7281-0858-2/19/\$31.00~\copyright~2019 IEEE. Personal use of this
        material is permitted. Permission from IEEE must be obtained for all other
        uses, in any current or future media, including reprinting\slash republishing
        this material for advertising or promotional purposes, creating new collective
        works, for resale or redistribution to servers or lists, or reuse of any
        copyrighted component of this work in other works. This material is referenced
        by DOI: {\href{https://doi.org/10.1109/BigData47090.2019.9005548}{10.1109/BigData47090.2019.9005548}}
    }
}

\begin{document}
\title{Learning and Recognizing Archeological Features from LiDAR Data}
\author{%
  \IEEEauthorblockN{Conrad M Albrecht}
  \IEEEauthorblockA{%
    \small
    \textit{TJ Watson Research Center, IBM Research}\\
    Yorktown Heights, NY 10598, USA\\
    \texttt{cmalbrec@us.ibm.com}
  }
  \and
  \IEEEauthorblockN{Chris Fisher}
  \IEEEauthorblockA{%
    \small
    \textit{Department of Antropology \& Geography}\\
    Colorado State University\\
    Fort Collins, CO 80523, USA\\
    \texttt{ctfisher@colostate.edu}
  }
  \and
  \IEEEauthorblockN{Marcus Freitag}
  \IEEEauthorblockA{%
    \small
    \textit{TJ Watson Research Center, IBM Research}\\
    Yorktown Heights, NY 10598, USA\\
    \texttt{mfreitag@us.ibm.com}
  }
  \and
  \IEEEauthorblockN{Hendrik F Hamann}
  \IEEEauthorblockA{%
    \small
    \textit{TJ Watson Research Center, IBM Research}\\
    Yorktown Heights, NY 10598, USA\\
    \texttt{hendrikh@us.ibm.com}
  }
  \and
  \IEEEauthorblockN{Sharathchandra Pankanti}
  \IEEEauthorblockA{%
    \small
    \textit{TJ Watson Research Center, IBM Research}\\
    Yorktown Heights, NY 10598, USA\\
    \texttt{sharat@us.ibm.com}
  }
  \and
  \IEEEauthorblockN{Florencia Pezzutti}
  \IEEEauthorblockA{%
    \small
    \textit{Department of Antropology \& Geography}\\
    Colorado State University\\
    Fort Collins, CO 80523, USA\\
    \texttt{fpezzutt@gmail.com}
  }
  \and
  \IEEEauthorblockN{\hphantom{phantom author}}
  \IEEEauthorblockA{\hphantom{phantom affiliation}}
  \and
  \IEEEauthorblockN{Francesca Rossi}
  \IEEEauthorblockA{%
    \small
    \textit{TJ Watson Research Center, IBM Research}\\
    Yorktown Heights, NY 10598, USA\\
    \texttt{francesca.rossi2@ibm.com}
  }
  \and
  \IEEEauthorblockN{\hphantom{ }}
  \IEEEauthorblockA{\hphantom{}}
}

\maketitle
\begin{abstract}
    We present a remote sensing pipeline that processes LiDAR
    (\textit{Li}ght \textit{D}etection \textit{A}nd \textit{R}anging)
    data through machine \& deep learning for the application of archeological
    feature detection on big geo-spatial data platforms such as e.g.\
    \textit{IBM PAIRS Geoscope} \cite{klein2015,lu2016}.

    Today, archeologists get overwhelmed by the task of visually surveying huge
    amounts of (raw) LiDAR data in order to identify areas of interest for inspection
    on the ground. We showcase a software system pipeline that results in significant
    savings in terms of expert productivity while missing only a small fraction
    of the artifacts.

    Our work employs artificial neural networks in conjunction with an efficient
    spatial segmentation procedure based on domain knowledge. Data processing is
    constraint by a limited amount of training labels and noisy LiDAR signals due
    to vegetation cover and decay of ancient structures. We aim at identifying
    geo-spatial areas with archeological artifacts in a supervised fashion allowing
    the domain expert to flexibly tune parameters based on her needs.
\end{abstract}
\thispagestyle{copyright}

\section{Introduction \& Motivation}

New remote sensing technologies such as LiDAR, cf.\ \cite{noaa2012} and references therein,
are revolutionizing many industries and fields---among them the one of archaeology---by
providing rapid, high resolution scans of topography which might reveal e.g.\ the
existence of ancient cities and landscapes. The problem is that, given the cost
and labor intensive nature of traditional methods, archaeologists cannot effectively
analyze these datasets. Via field-work and manual mapping, they exploit
their domain knowledge to recognize human artifacts and classify them as houses,
temples, walls, streets, and other elements of past human settlements.

LiDAR has the capacity to scan and map hundreds of square kilometers in a significantly
shorter time compared to traditional archaeological fieldwork. These LiDAR datasets
are delivered as a cloud of points with 3D information. The task of manual and
computer-aided extraction and mapping of ground features for analytical purposes
is a significant challenge and has attracted attention in the archeological
and remote sensing literature since about the mid-2000s: \cite{
  bewley2005, harmon2006, doneus2008, riley2009, coluzzi2010, trierpilo2012,
  kyamme2013, luo2014, fisher2016, hazell2017, canuto2018, guyot2018, tarolli2019%
}.

In this paper we describe how techniques from computer vision and machine learning
provide pipelines implemented on top of geospatial big data platforms such as
e.g.\ \textit{IBM PAIRS Geoscope}\footnote{%
    access via \href{https://ibmpairs.mybluemix.net}{https://ibmpairs.mybluemix.net},
    open-source Python API wrapper available at
    \href{https://pypi.org/project/ibmpairs}{https://pypi.org/project/ibmpairs}
    with code development in GitHub:
    \href{https://github.com/IBM/ibmpairs}{https://github.com/IBM/ibmpairs},
    recently an Anaconda package has been published as well:
    \href{https://anaconda.org/conda-forge/ibmpairs}{https://anaconda.org/conda-forge/ibmpairs}
} to accelerate the archeologist's work. Using LiDAR data from the ancient
Purepecha city of Angamuco in Mexico \cite{chase2012}, we exploit domain knowledge
to learn and recognize promising ancient artifacts in this city such as e.g. houses.
This allows us to more effectively identify areas of interest to the archeologists
and to automatically classify and localize ancient, human-made artifacts.

We filter the data according to the archeologists' domain knowledge on size, shape,
and similar features. We then pass these shapes---in
the form of an image bounded by an almost minimum bounding box---to a machine learning
classifier that recognizes if they are human artifacts. The classifier follows an
ensemble methodology based on a set of trained VGG artificial network \cite{simonyan2014very}
using manually annotated images given by the archeologists. The experimental results
show that this approach is accurate and flexible for the archeologist's needs.
\begin{figure}[t]
  \centering
  \includegraphics[width=\columnwidth]{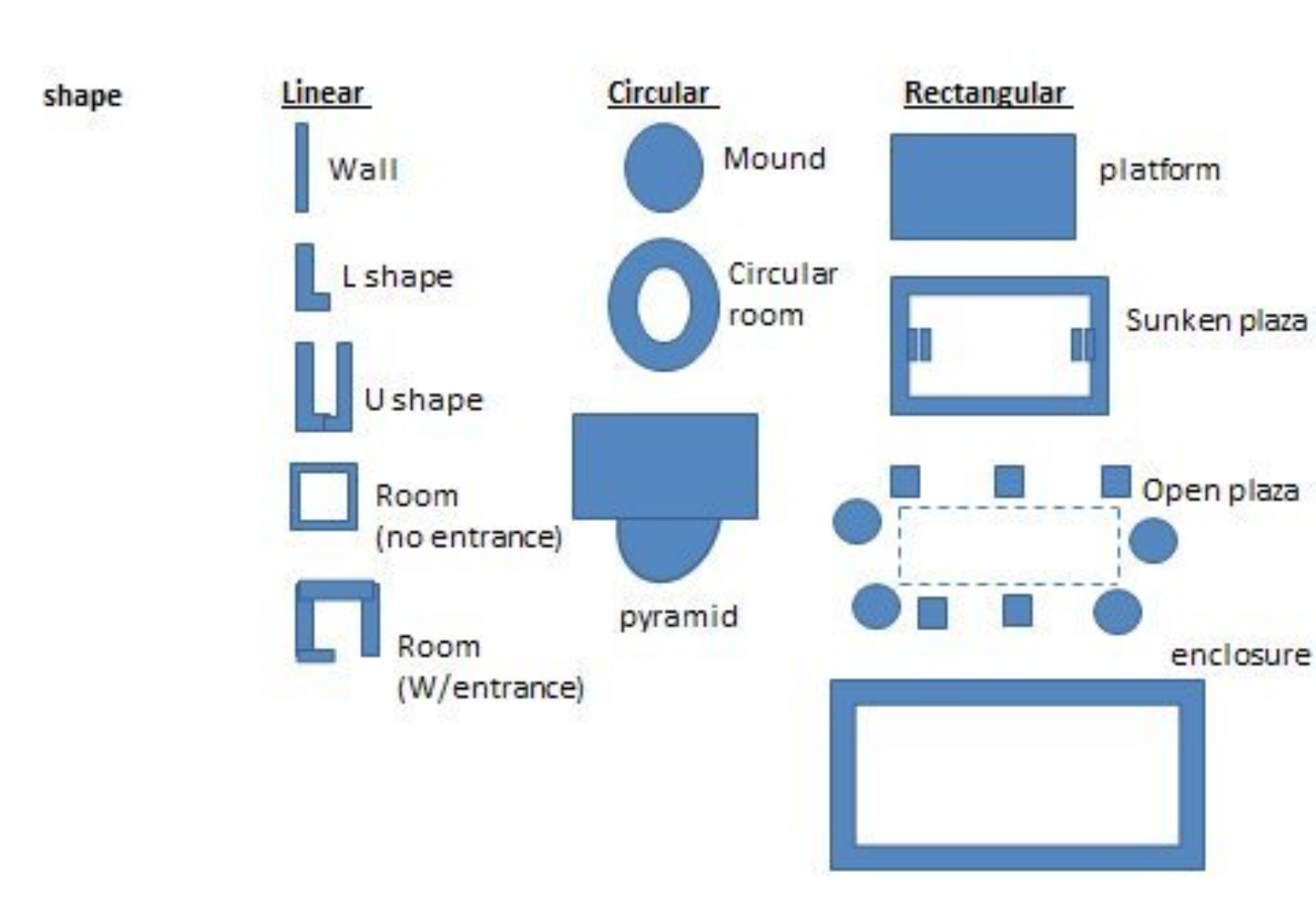}
  \caption{\label{fig:domain}
      Overview of archeological classification scheme wrt.\ ancient buildings.
  }
\end{figure}

\section{Domain Knowledge}

We consider data from the ancient city of Angamuco, Mexico,
where the archeologists already performed extensive fieldwork, manually recognizing and mapping
local features into specific types of human-made artifacts: houses, walls, temples, pyramids, etc.
We use these labels to train and test deep learning classifiers. We also exploit the archeologists'
domain knowledge of this area and of the ancient Purepecha civilization which dominated Western
Mexico during the last few centuries prior to European Conquest to guide us in our process of automating
feature extraction and recognition from LiDAR data.

The following is a brief description of the kinds of structures present in the area,
summarized in Fig.~\ref{fig:domain}.
\begin{itemize}
\item
    Linear features consist of single \brc{e.g. walls or roads} that are longer than wide,
    and their height consists of two rock courses---which translate on average to
    0.4--0.5m in height. Double linear features consists usually of more than one
    linear feature aggregated into a structure. For example, an ``L'' shape building
    consists of two linear features which creates an internal space\slash room on
    the inside.
    Triple linear features consists mostly of buildings walls \brc{three walls}
    with an entrance \brc{e.g., a house with entrance or a ``U'' building}. Finally,
    a structure composed of four linear features \brc{with no entrance} consists
    of a room.
\item
    Circular features can consist of mounds or circular rooms \brc{that is, a linear
    feature with a circular geometry}. Sometimes, pyramids can contain a circular
    or semi--circular sub--element \brc{mostly to the front of feature}. These
    features are most often placed facing a plaza \brc{sunken or open}.
\item
    Rectangular features consist of platforms \brc{with well-defined corners},
    sunken plazas, and open plazas \brc{conformed by other features} defined by
    the empty space formalized by such features.
\item
    Square features consist of rooms \brc{with and without entrance} and enclosures,
    which are of same shape as a room, but larger in area, enclosing a space.
\end{itemize}

\begin{figure}[t]
  \centering
  \includegraphics[width=.8\columnwidth]{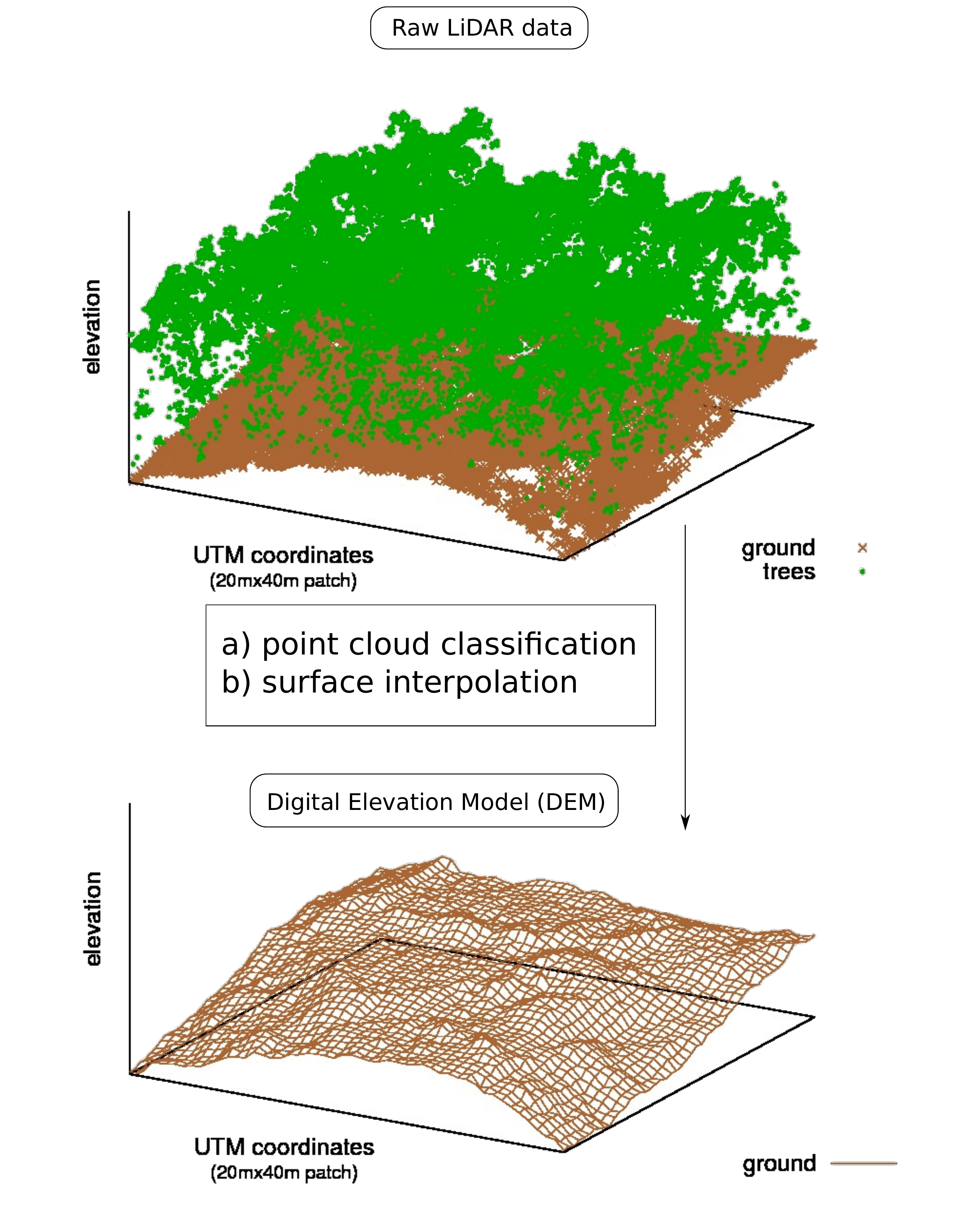}
  \caption{\label{fig:FromRawLiDARDataToDEM}
      Classified LiDAR point cloud and derived digital elevation model (DEM).
  }
\end{figure}

\section{Data Segmentation: From LiDAR Measurements to Images}
\label{sec:ImageSamplingFromLiDARData}
The raw data used to start our analysis is geo-referenced elevation points derived
from the physical LiDAR measurements
\begin{align}
    X_{ij} \equiv (x_i,y_i,z_{ij})
\end{align}
where $x_i$ and $y_i$ represent geo-location information, cf.\ longitude and latitude
\cite{maling2013}. For each coordinate pair $(x_i,y_i)$ with fixed $i$ there might exist many
elevation measurements $z_{ij}$ due to multiple returns of the LiDAR laser pulse from e.g.\
vegetation. In contrast, for solid surfaces, such as e.g.\ streets, there is a unique $z_i=z_{i0}$.
Since other surfaces such as e.g.\ water strongly absorb laser light, there might not even
exist a single $z_i$. Hence, the $(x_i, y_i)$ points form an irregular grid of data points.

\begin{figure*}[t]
    \centering
    \includegraphics[width=.45\textwidth]{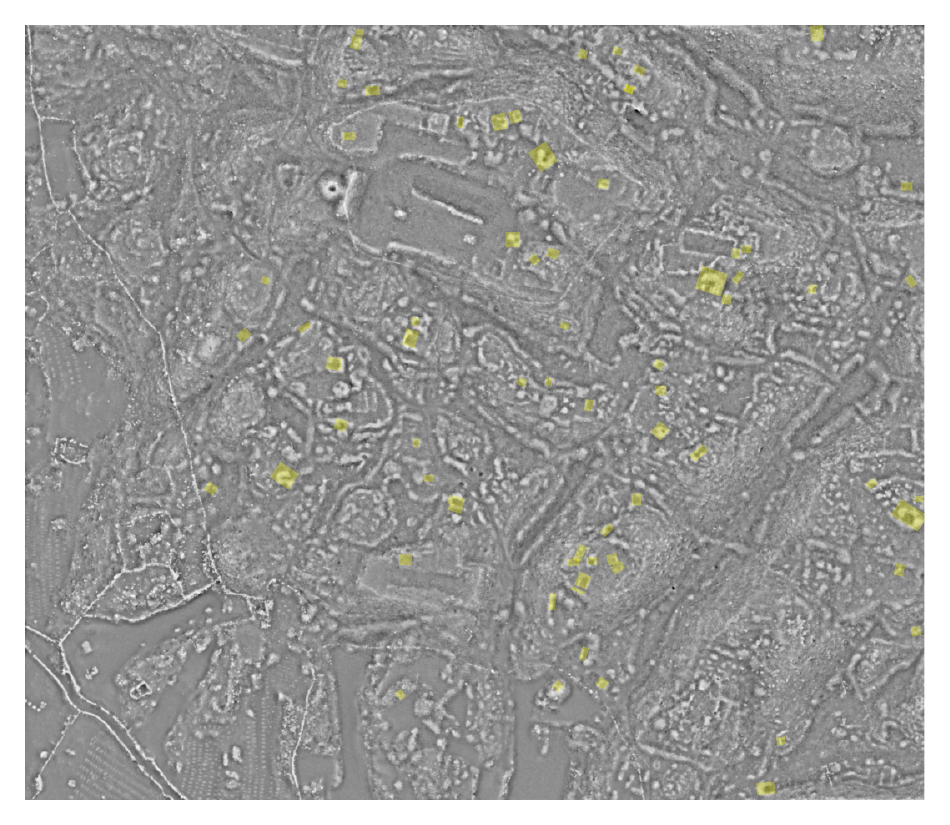}
    \includegraphics[width=.45\textwidth]{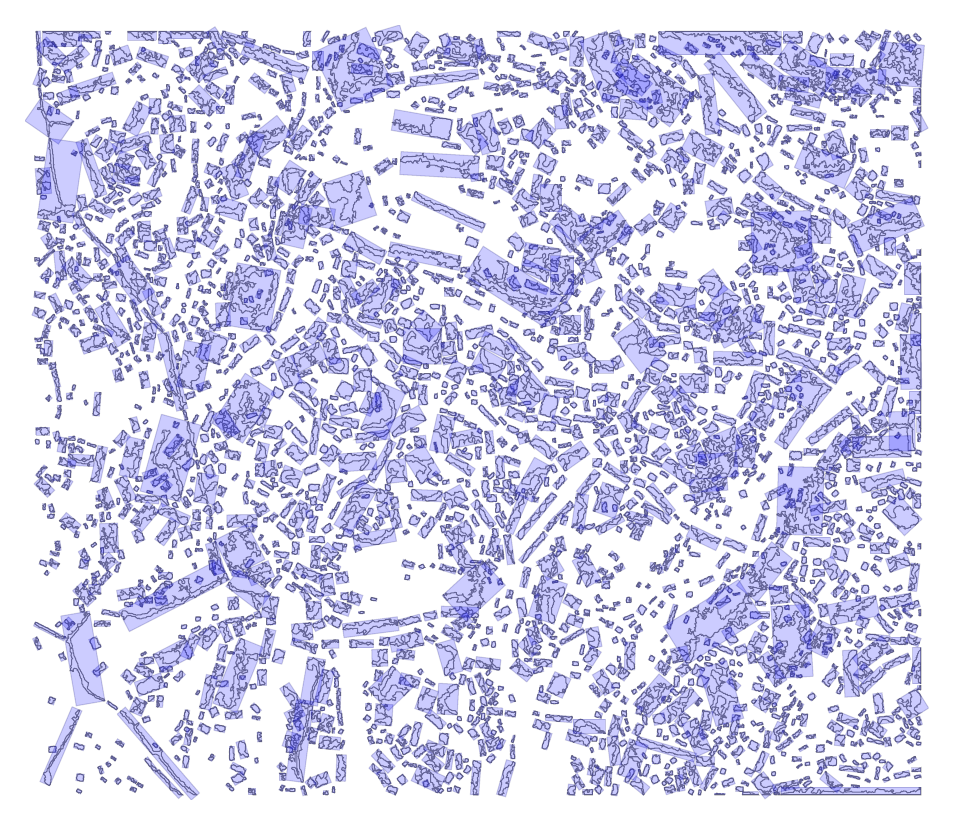}
    \caption{\label{fig:FromLocalDEMToMBBs}
        Left: the local DEM we derive, including manual annotations \brc{yellow}
        of human made structures that occupy an area equal to or larger than $20m^2$.
        Right: the contours \brc{black} we employ to define the minimum bounding
        boxes \brc{MBBs, blue} for image cropping. Those images are then fed
        to the deep learning classifier. The corresponding labeling procedure is
        depicted in Fig.\ \ref{fig:BoundingBoxClassification}.
    }
\end{figure*}
In a first step we classify \textit{bare ground} data points, representing the earth's surface or
 the top of a foundation, $X_{i}$ in order to interpolate them to a regular grid, a
 \textit{Digital Elevation Model} (DEM). We apply the simple transformation
\begin{align}
    X_{ij} \to X_i \equiv (x_i, y_i, z_{i0})
\end{align}
with $z_{ij}<z_{ij'}\quad\forall~ i,j<j'$. A nearest neighbor interpolation
is used to convert the $X_i$ to a regular grid, 
with a spatial resolution of about 0.3 meters\footnote{This roughly corresponds to the
average density of the LiDAR data scan.}:
\begin{align}
    X_i\to X(x,y)
    \quad,
\end{align}
see Fig.\ \ref{fig:FromRawLiDARDataToDEM}.

Our approach does not filter outliers due to the fact that there exist series $z_{ij}$
where the \textit{last return} $z_{i0}$ does not represent bare ground. As we will
see, our processing pipeline applies thresholding that naturally cuts off those peaks. To
improve the image quality one could apply classical image filters as e.g.\ a linear
Gaussian or a non-linear Median filter \cite{buades2005}.
In contrast our simplified strategy benefits from increased speed compared to more
elaborate DEM generation techniques.

Since we are interested in cultural features such as e.g.\ houses, only, and the irregular terrain dominates
the DEM, we filter our global DEM $X(x,y)$ suppressing wavelengths above a certain length scale $\Lambda$
by means of a two-dimensional Fourier analysis \cite{ersoy1994}.
~$\Lambda$ defines our notion of \textit{local}. For our experiments we use $\Lambda$ in the
range of couple of meters ($\sim10$ pixels). The parameter is manually tuned to
yield satisfactory results by visual inspection. The left part of Fig.\ \ref{fig:FromLocalDEMToMBBs}
provides an example.

\begin{figure}[t]
  \centering
  \includegraphics[width=.8\linewidth]{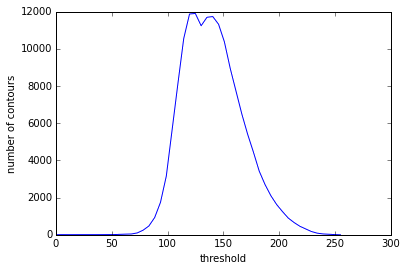}
  \caption{\label{fig:ThresholdTuning}
  Number of contours $\vert\bar{\mathfrak{C}}\vert$ for given threshold parameter $\Delta_0$.
  Values on the abscissa encode the normalized, one-byte grayscale values of the local
  DEM image on the left-hand side of Fig.\ \ref{fig:FromLocalDEMToMBBs}.
  Automatically setting $\Delta_0$ amounts for maximizing the number of MBBs and
  their associated contours, respectively. While $\Delta_0\sim 0$ 
  typically results in a single MBB covering the total survey area,
  it gets partitioned into multiple smaller MBBs for increasing $\Delta_0$ to the
  point where MBBs shrink to zero in size until none is left for $\Delta_0=255$.
  Depending on the complexity of the local DEM, the number of MBBs will typically
  fluctuate for $\Delta_0$ near $256/2=128$: While some MBBs shrink to zero area,
  others might partition into smaller ones.
  }
\end{figure}

Based on this \textit{local} DEM $\tilde  X(x,y)$ we extract a set of contours $\mathfrak{C}$ 
for a fixed elevation $\Delta_0$:
\begin{align}
    \mathfrak{C}\equiv\{\mathcal{C}_k:\tilde X(x,y)=\Delta_0\}
\end{align}
which we again manually tune in a physically
reasonable range of about $0.2$ to $0.5$ meters above local\slash reference ground
$\Delta_0=0$. Of course, this procedure can be automatized by optimization as detailed
in Fig.\ \ref{fig:ThresholdTuning}.

Accounting for the contour's hierarchy we reduce the set of contours
to
\begin{align}
    \bar{\mathfrak{C}}\equiv
    \{
        \mathcal{C}_k:
        C_k\cap C_{k'}\neq\emptyset\Rightarrow C_{k'}\subset C_k
        ~\forall\mathcal{C}_{k'}\in\mathfrak{C}
    \}
\end{align}
where $C_k$ denotes the area enclosed by $\mathcal{C}_k$, i.e.\ $\partial C_k=\mathcal{C}_k$.
Now we derive the set of \textit{Minimum Bounding Boxes} (MBBs) \cite{freeman1975}:
\begin{align}
    \bar{\mathfrak{R}}\equiv
    \{
        \mathcal{R}_k:
        A(\mathcal{R}_k)\leq A(\mathcal{R})
        ~\forall\mathcal{R}\supset\mathcal{C}_k\in\bar{\mathfrak{C}}
    \}
\end{align}
where $\mathcal{R}_k$ defines a rectangle and $A(\mathcal{R}_k)$ its area.
Fig.\ \ref{fig:FromLocalDEMToMBBs} (right part) shows both: $\bar{\mathfrak{C}}$
\brc{black contours} and $\bar{\mathfrak{R}}$ \brc{blue, semi-transparent boxes}.

Note that $\bar{\mathfrak{R}}$ defines relevant areas of interest to crop images
from $\tilde X_i$. This approach allows us to be more efficient compared to a naive
\textit{sliding window} procedure where one needs to systematically scan trough all
possible rectangular window sizes shifted through the whole two-dimensional area!

Moreover, the MBBs allow us to apply pre-filtering to discard noise or irrelevant
contours. In particular, since we are interested in recognizing ruins corresponding
to house-like structures, we further restrict $\bar{\mathfrak{R}}$ such that all
$\mathcal{R}_k$ have an area of at least $3m^2$, an aspect ratio of less than $1:10$,
and a circumference within bounds of 10 to 200 meters---since this is the typical
size of a house in that area, according to the archeologists. In Fig.\ \ref{fig:TestTrainingDataFlow}
we label the processing from $\tilde X(x,y)$ to $\bar{\mathfrak{R}}$ including pre-filtering
by\circled{P1}. For our specific test setting we have $\vert\bar{\mathfrak{R}}\vert=1805$.

\section{Deep Learning: From Images to House Classification}
\label{sec:DeepLearningAndPerformanceBenchmarking}

Starting from the MBBs $\bar{\mathfrak{R}}$ generated as outlined in the previous section,
we crop $\tilde X(x,y)$ to obtain the
image set
\begin{align}
    \mathfrak{I}\equiv
    \{
        I_k(x,y):
        \tilde{X}(x,y)\subseteq\mathcal{R}_k
        ~\forall\mathcal{R}_k\in\bar{\mathfrak{R}}
    \}
    \quad.
\end{align}
Let us denote the house classification function as
\begin{align}
    h(\mathcal{Y})\equiv
    \begin{cases}
        1&\mathcal{Y}\text{ contains house ruins}\\
        0&\text{else}
    \end{cases}
\end{align}
with $\mathcal{Y}$ a set of (geo-referenced) polygons. We consider the set of manual annotations
of house ruins from the archeological field survey:
\begin{align}
    \mathfrak{H}\equiv
    \{
        \mathcal{H}_k:
        h(\mathcal{H}_k)=1
        \land
        A(\mathcal{H}_k)\geq20m^2
    \}
    \quad,
\end{align}
i.e.\ for our case study we pick archeological ruins representing house-like structures
which occupy an area greater or equal to 20 square meters. For our benchmark we
have $\vert\mathfrak{H}\vert=70$.

In order to classify the elements of $\mathfrak{I}$ we particularly use
\begin{align}
    \label{eq:HouseClassificationFunction}
    h(\mathcal{R}_k)\equiv
    \begin{cases}
        1&
        A(R_k\cap\mathcal{H}_k)/A(R_k)\geq a_1\\
        &\land~
        A(R_k\cap\mathcal{H}_k)/A(\mathcal{H}_k)\geq a_2\\
        0&\text{else}
    \end{cases}
\end{align}
where $\mathcal{R}_k$ is the MBB corresponding to the image $I_k$ and $\mathcal{H}_k$
references any manual house annotation $\mathcal{H}_k\in\mathfrak{H}$ with $\mathcal{R}_k\cap\mathcal{H}_k\neq\emptyset$.
For our evaluation we set the constants to
\begin{align}
    a_1 = a_2 = 0.3
    \quad.
\end{align}
\begin{figure}[t]
  \centering
  \includegraphics[width=.7\linewidth]{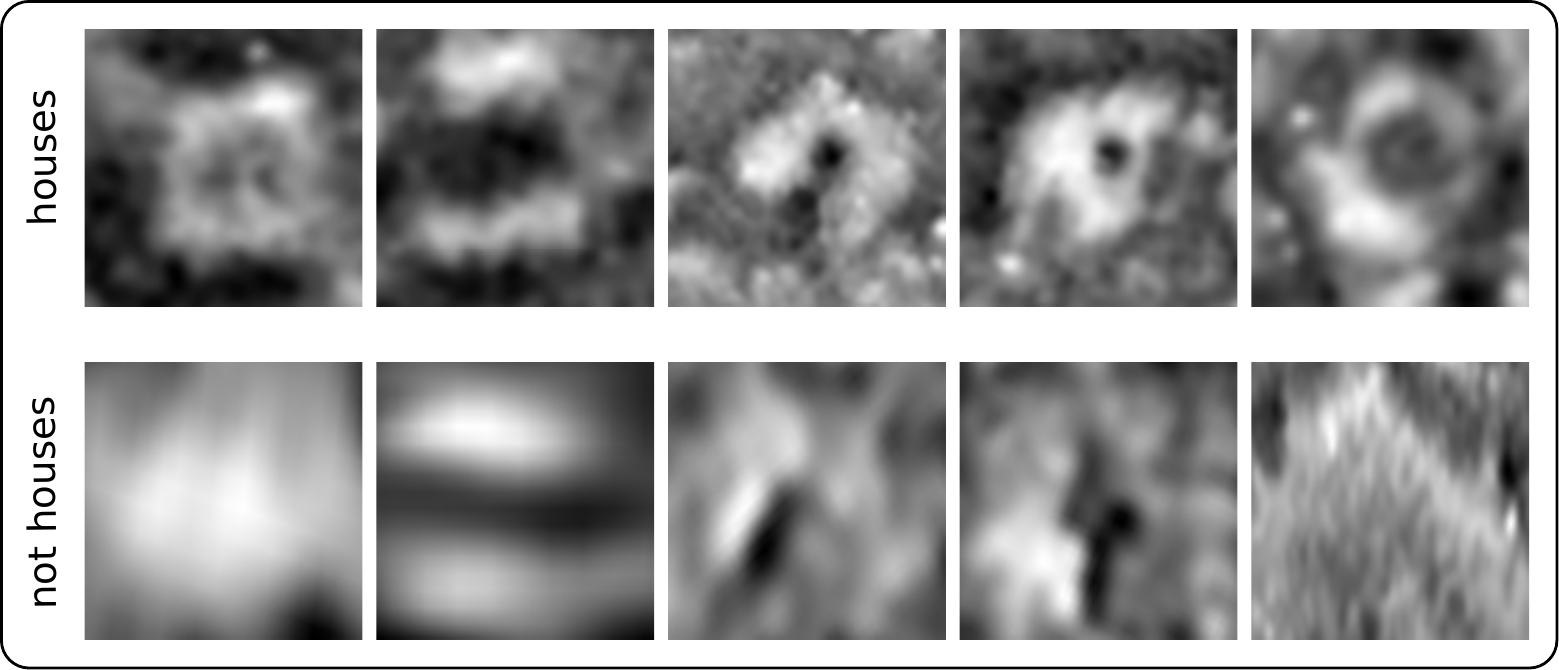}
  \caption{\label{fig:MachineLearningPictureSamples}
      Sample images $J_k\in\mathfrak{J}$ \brc{$100\times100$ pixels} for areas
      with ruins classified as ``houses'' by the archeologists \brc{top row}. Negative
      samples are shown at the bottom row.
  }
\end{figure}

To increase the number $\vert\mathfrak{I}\vert$ of images as well as including the
house feature's \textit{context}, we multiply the set $\bar{\mathfrak{R}}$ by
expanding each $\mathcal{R}_k\in\bar{\mathfrak{R}}$ such that it includes up to 2
meters of $\mathcal{R}_k$'s surroundings in steps of $\tfrac{1}{3}m$. We perform
the increase of $A(\mathcal{R}_k)$ by a parallel shift of the boundary $\partial\mathcal{R}_k$.

The feature vector we supply to the deep learning algorithm is constructed from
an affine transformation $f_N(I)$ of the $I_k(x,y)$ to the space
\begin{align}
    \mathbb{N}^2_N\equiv
    \mathbb{N}\cap(0,N]~\times~\mathbb{N}\cap(0,N]
    \quad,
\end{align}
i.e.\ an image with aspect ratio 1 and a number of $\vert\mathbb{N}^2_N\vert=N^2$
pixels. Our experiments fix $N=100$ to obtain the feature vectors
\begin{align}
    \mathfrak{I}\to
    \mathfrak{J}\equiv
    \{
        J_{k,ij}=\left(f_N(I_k)\right)_{ij}\in\mathbb{N}^2_{100}
    \}
    \quad.
\end{align}
\begin{figure}[t]
  \centering
  \includegraphics[width=.7\linewidth]{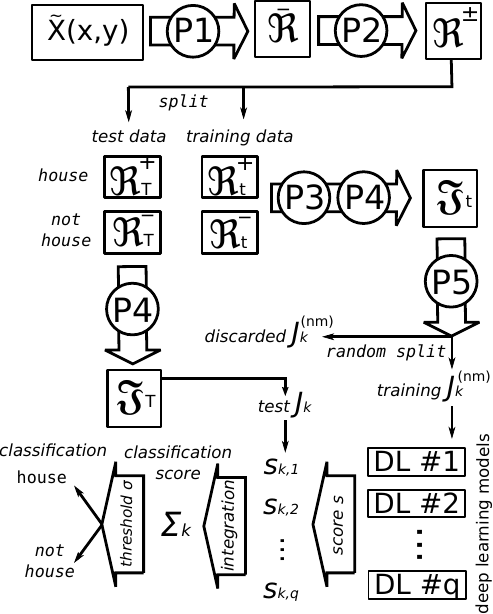}
  \caption{\label{fig:TestTrainingDataFlow}
      Flow chart for generating and splitting test and training data of our
      experimental setup for identifying house-like structures from LiDAR data.
      The circled process labels P1\dots5 are detailed in the main text.
  }
\end{figure}
\begin{figure*}[t]
  \centering
  \includegraphics[width=.8\linewidth]{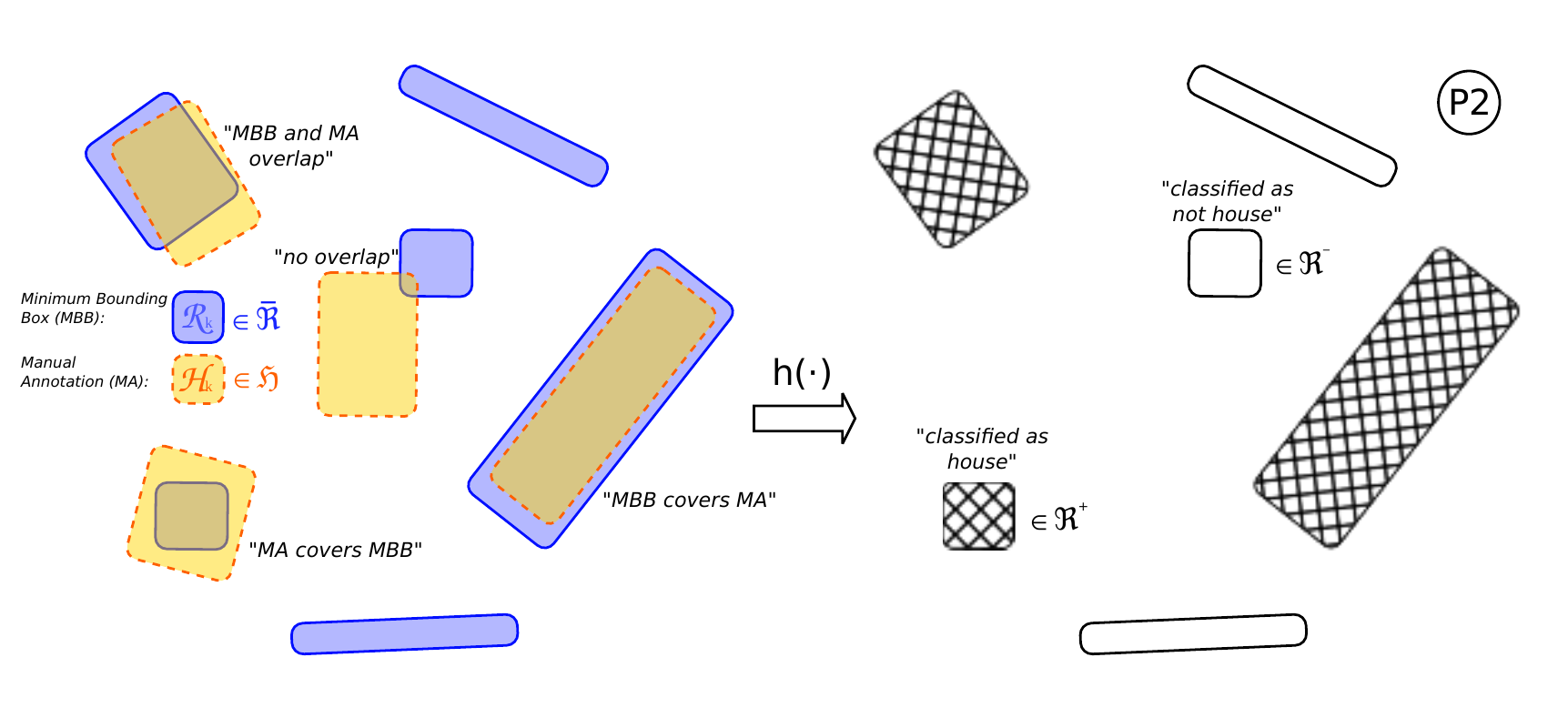}
  \caption{\label{fig:BoundingBoxClassification}
     Cartoon to illustrate the classification of $\mathfrak{R}^{\pm}$ from the local
     DEM cropped by the MBBs \brc{blue, solid boundary} from manual annotation of
     domain experts \brc{yellow, dashed boundary}.
  }
\end{figure*}

Moreover, we apply a \textit{normalization} to each $J_{k,ij}\in\mathfrak{J}$ according
to
\begin{align}
    J_{k,ij}\to
    \frac{J_{k,ij}-\langle J_k\rangle}{\max_{i,j}J_{k,ij}-\min_{i,j}J_{k,ij}}
\end{align}
such that the normalized images have vanishing mean $\langle J_k\rangle=0$ and the
absolute height (to global terrain) gets scaled out. Here, $\langle\cdot\rangle$
denotes averaging over image pixels, i.e.
\begin{align}
    \langle J_k\rangle=\frac{1}{N^2}\sum_{i,j}J_{k,ij}
\end{align}
We present a collection of sample images $J_k$ of house-like structures, as well
as areas without houses, in Fig.\ \ref{fig:MachineLearningPictureSamples}. The full
process of generating $J_k\in\mathfrak{J}$ from MBBs $\mathcal{R}_k\in\bar{\mathfrak{R}}$
is labeled as\circled{P4}in Fig.\ \ref{fig:TestTrainingDataFlow}.

In order to a) increase the number $\vert\mathfrak{J}\vert$ of images for training
the deep learning algorithm and b) to take into account the feature's \textit{context},
we multiply the number of elements of the set $\bar{\mathfrak{R}}$ by expanding each
$\mathcal{R}_k\in\bar{\mathfrak{R}}$ such that it includes up to,
but less than 2 meters of $\mathcal{R}_k$'s surroundings in steps of $\tfrac{1}{3}m$.
We perform the increase of $A(\mathcal{R}_k)$ by an \textit{outward} parallel shift
of the boundary faces $\partial\mathcal{R}_k$. Specifically, for each MBB $\mathcal{R}_k\in\bar{\mathfrak{R}}$
we generate 6 \textit{widened} MBBs $\mathcal{R}_k^{0m}$, $\mathcal{R}_k^{\tfrac{1}{3}}$,
\dots $\mathcal{R}_k^{1\tfrac{2}{3}m}$ to obtain corresponding normalized images $J^{(n)}_k$,
$n=0,\dots,5$. We label this process by\circled{P3}. Another multiplication
factor of 4 is achieved by rotating each $J^{(n)}_k$ by angles of $\pi/2$:
\begin{align}
    J^{(n)}_{k,ij}\to~&
    J^{(n0)}_{k,ij}\equiv J^{(n)}_{k,ij}~,~~
    J^{(n1)}_{k,ij}\equiv J^{(n)}_{k,jN-i}~,\nonumber\\
    &J^{(n2)}_{k,ij}\equiv J^{(n)}_{k,N-ji}~,~~
    J^{(n3)}_{k,ij}\equiv J^{(n)}_{k,N-iN-j}
    \quad.
\end{align}
which we refer to as\circled{P5}.

Finally, for our performance benchmarking we take $\bar{\mathfrak{R}}$ and apply the
classification function $h$, cf.\ Eq.\ \ref{eq:HouseClassificationFunction}, to define
\begin{align}
    \mathfrak{R}^+&\equiv
    \{
        \mathcal{R}_k:
        h(\mathcal{R}_k)=1
        ~\forall\mathcal{R}_k\in\bar{\mathfrak{R}}
    \}\\
    \mathfrak{R}^-&\equiv
    \{
        \mathcal{R}_k:
        h(\mathcal{R}_k)=0
        ~\forall\mathcal{R}_k\in\bar{\mathfrak{R}}
    \}
\end{align}
such that $\bar{\mathfrak{R}}=\mathfrak{R}^+\cup\mathfrak{R}^-$, cf.\
process\circled{P2}depicted in Fig.\ \ref{fig:BoundingBoxClassification}.
Then, we randomly split $\mathfrak{R}^\pm$ into a training and a test set:
$\mathfrak{R}^\pm_t$ and $\mathfrak{R}^\pm_T$, respectively, such that
$\mathfrak{R}^+_T\cup\mathfrak{R}^+_t=\mathfrak{R}^+$ and
$\mathfrak{R}^-_T\cup\mathfrak{R}^-_t=\mathfrak{R}^-$.
In our specific setting we have
\begin{center}
    \begin{tabular}{c|cccc}
        &
        $\mathfrak{R}^+_t$&
        $\mathfrak{R}^-_t$&
        $\mathfrak{R}^+_T$&
        $\mathfrak{R}^-_T$\\
        \hline
        $\vert\cdot\vert$&
        44&
        1056&
        16&
        689
    \end{tabular}
\end{center}
Note that $\vert\bar{\mathfrak{R}}^+\vert/\vert\mathfrak{H}\vert\approx0.86<1$,
i.e.\ the MBBs and\slash or LiDAR data do not perfectly capture all signatures of houses
surveyed by the archeologists in the area of interest. There exists artifacts
labeled that are not represented by an appropriate MBB derived from the local DEM
which can be traced back to the fact that the corresponding local wall structures
eroded below the picked threshold $\Delta_0$. Another root cause is heavy vegetation
cover that does not allow for sufficient many LiDAR pulses to reach bare ground.

Starting from a local DEM, the general processing pipeline generates test and training
data to be fed as input to models using {\em 64} fully connected {\em fc} deeply learnt VGG network
representation feature set \cite{simonyan2014very}. The workflow\footnote{
    However, since in our case $\vert\mathfrak{R}_t^-\vert\gg\vert\mathfrak{R}_t^+\vert$,
    we apply only\circled{P4} to the elements of $\mathfrak{R}_t^-$. In fact this
    leads to an equal number of positive and negative training samples: $44\cdot6\cdot4=1056$.
} is depicted in Fig. \ref{fig:TestTrainingDataFlow}.
\begin{figure*}[t]
    \begin{subfigure}[b]{.48\linewidth}
        \centering
        \includegraphics[width=1.1\textwidth]{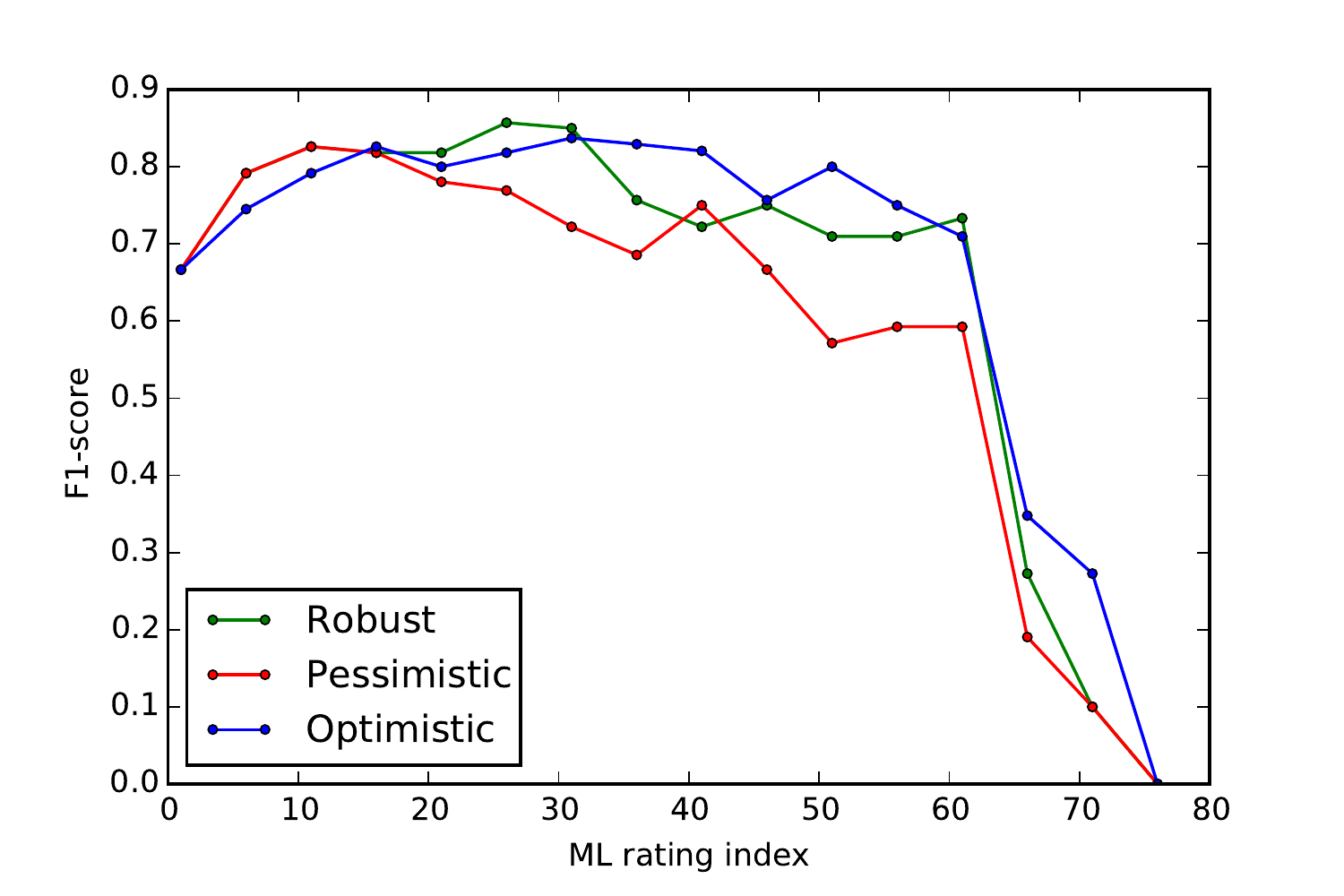}
        \caption{
            $F1$-score, that is the harmonic mean of precision and recall \cite{davis2006},
            for house detection, plotted vs.\ the \textit{ML rating index} threshold $100\sigma$.
        }
    \end{subfigure}\hfill
    \begin{subfigure}[b]{.48\linewidth}
        \centering
        \includegraphics[width=1.1\textwidth]{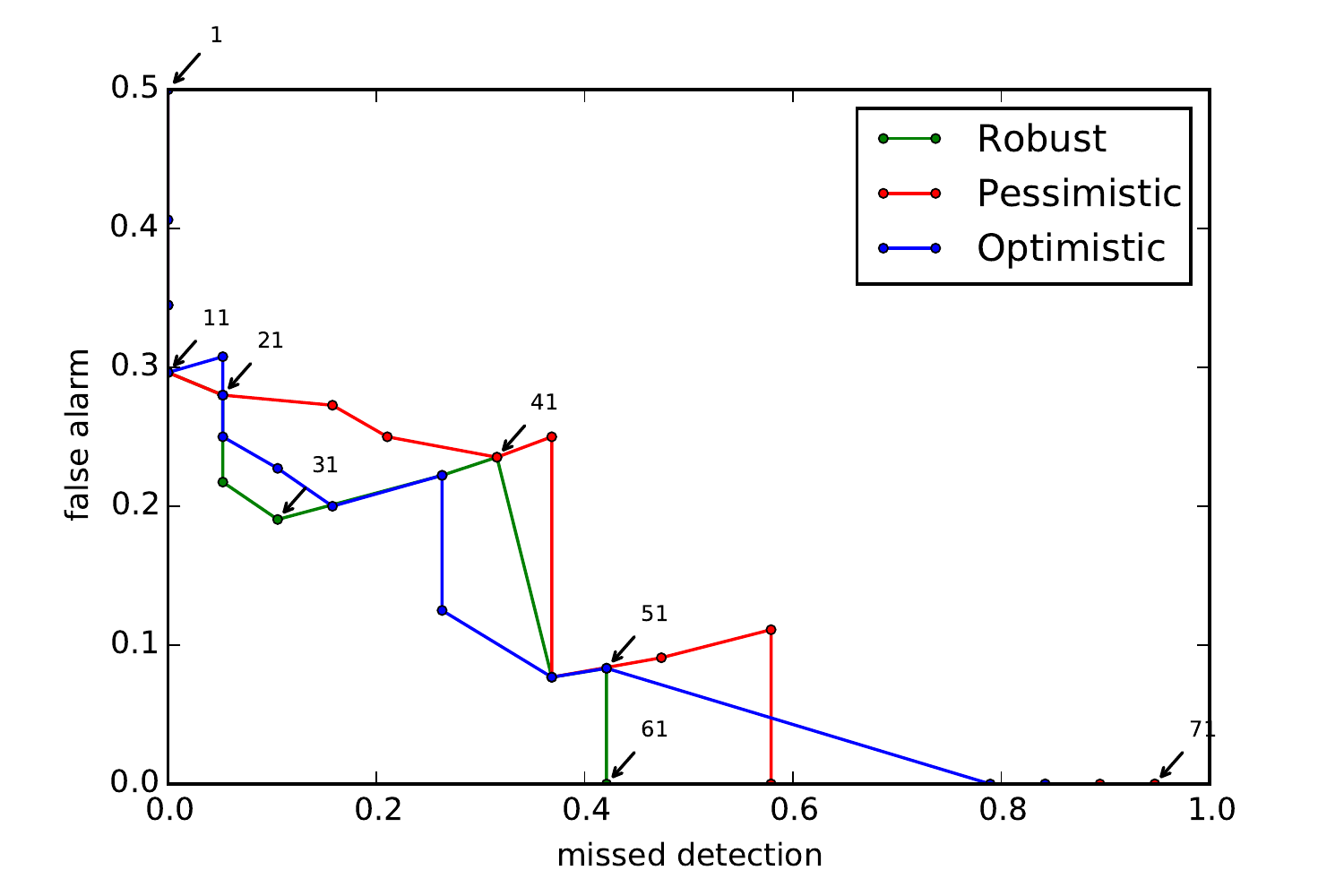}
        \caption{
            The Detection Error Tradeoff (DET) graph is obtained by tuning the
            \textit{ML rating index threshold} \brc{black arrows with label}.
        }
    \end{subfigure}
    \caption{\label{fig:PerformanceAnalysis}
        Numerical evaluation of house recognition performance. The three different
        curves \brc{red, blue, green} correspond to different schemes how elementary
        classifiers get combined, for details cf.\ main text.
    }
\end{figure*}

Taking the $J_k^{(nm)}\in\mathfrak{J}_t$ we randomly select {\em 90\%} of the
training sample data into a sample training run. Ten such random selections
\brc{$q=10$ in Fig.\ \ref{fig:TestTrainingDataFlow}} result in ten different training
runs and corresponding house models. Each machine learnt model \textsl{DL \#1},
\textsl{DL \#2},\dots,\textsl{DL \#10} defines a classifier, each returning a confidence
score $s_k\in[0,1]$. A lower score indicates that the input image is less likely
to be a house and the higher score suggests that the input image is more likely
to be a house.

An integration function $\Sigma$ generates a classification score $\Sigma_k$ for
a given test image $J_k$. In particular we employ the median, minimum, and maximum
of the scores $s_{k,1}$, $s_{k,2}$, \dots, $s_{k,10}$ to define three three fused classifier
models: \textit{robust}, \textit{pessimistic}, and \textit{optimistic}, respectively.

We use a variable threshold $\sigma\in[0,1]$ for classification according to:
$\Sigma_k>\sigma\Rightarrow J_k~\text{\tt is house}$, and $J_k$ \texttt{not a house},
elsewise. The numerical analysis of our pipeline's performance is shown for each
of the fused classifier models in Fig.~\ref{fig:PerformanceAnalysis} in terms of
the $F1$--score as well as by plotting the detection error tradeoff (DET).

\section{Discussion of Results \& Perspective}

We are encouraged by our findings: As highlighted in the abstract, cognitive analytics
can result in significant savings in terms of expert productivity\footnote{
    The manual annotations used in $\mathfrak{H}$ have been collected as part of
    several surveys that multiple archeologists collected on the order of years while
    computation was performed on the order of hours.
} while missing a fraction of the artifacts---if permissible by the application.

The primary motivation of our remote sensing pipeline for archeology springs from
the need of scaling the archeologist's expertise. It is simply infeasible to allocate
larger number of experts or taking longer periods of time to mark every structure 
by the archeologist meticulously looking for tell-tale signs distinguishing deteriorating
evidence. Consequently, the existing best practices often suggest marking a few
randomly selected artifacts in large field being surveyed and relying on using
statistical techniques for estimating the number of artifacts in the entire field.

After a careful analysis of the available data and repertoire of the artifacts
present within the data, cf.\ Fig. \ref{fig:domain}, we chose {\em house} as
a representative artifact to assess the efficacy of the cognitive approach to scale
archaeological expertise: These artifacts can be both high in volume, small in spatial
extent, and could potentially be very similar to other structures such as {\it round house}.

Following our first experiments on a moderate size LiDAR dataset, in order to draw
broad conclusions, our study aims at expanding the data processing to larger archeological
sites, in particular with the aid of the Big Geospatial Data platform \textit{IBM PAIRS}.

Secondly, our methodology allowed us to randomly distribute both test and training
samples both originating from the same geographic area. In this respect, the reported
performance of our system is an optimistic estimate with the expectation that the
performance will generalize to the artifact feature population in the extended geographic
area of Western Mexico.

Finally, our system processing constrained all the \brc{cropped} images to be resized
to the same dimensions and be thus agnostic to the scale information while any real
system would be able to leverage scale information for recognizing the patterns.
However, we aim at employing MBB characteristics such as, e.g., area $A(\mathcal{R}_k)$,
circumference, aspect ratio, $A(\mathcal{C}_k)/A(\mathcal{R}_k)$, and number of contours
at fixed $\Delta_0$ in $\mathcal{R}_k$, to be either directly incorporated into
the feature vector of $J_k$ or to be fed to a separate machine learning model such
as random forest. Table \ \ref{tab:featureClassifiers} shows actual results of the
latter approach that can be merged with the classification of the deep learning
models \textsl{DL \#1\dots q}.

To add to the qualitative analysis, we observe the following:
As we can see from in Fig.~\ref{fig:PerformanceAnalysis}, the fused classifiers are
performing at equal error rate (EER) in the range of  {\em 0.22} to {\em 0.25}. 
Depending upon the application needs and the availability of the experts, the operating
point of the system can be adjusted to scale the operation of finding archaeological
artifacts. For example, if the solution requires very accurate estimates of the
detected artifacts, the operating point of the system needs to be shifted to left
\brc{e.g., to small missed detection rate} leading to relatively large false alarm
rate and thus requiring more expert time to sieve through the real detections from
the false alarms.
On the other hand, if the solution can accept approximate answers, the system can
be operated on the right hand side of the DET curve which will result in relatively
fewer false alarms, i.e., less human oversight needed, at the risk of missing
genuine artifacts.
\begin{table*}[h!]
    \relsize{-0}
\centering
\begin{center}
\begin{tabular}{lllllllll}
\bf classifier &\bf TP rate &\bf FP rate &\bf precision &\bf recall &\bf F--measure &\bf MCC &\bf ROC area\\
\hline\hline
\tt lazy-kstar                  & 0.650 & 0.191 & 0.972 & 0.650 & 0.767 & 0.142 & 0.825 \\ \hline
\tt meta-Logitboost             & 0.706 & 0.190 & 0.973 & 0.706 & 0.808 & 0.167 & 0.807 \\ \hline
\tt MetaAdaboostM1              & 0.779 & 0.310 & 0.970 & 0.779 & 0.856 & 0.166 & 0.800 \\ \hline
\tt functions-Logistic          & 0.706 & 0.251 & 0.971 & 0.706 & 0.808 & 0.147 & 0.794 \\ \hline
\tt lazy-LWL                    & 0.465 & 0.073 & 0.975 & 0.465 & 0.612 & 0.117 & 0.793 \\ \hline
\tt functions-SimpleLogistic    & 0.709 & 0.312 & 0.969 & 0.709 & 0.810 & 0.129 & 0.790 \\ \hline
\tt trees-randomforest          & 0.573 & 0.193 & 0.971 & 0.573 & 0.707 & 0.114 & 0.784 \\ \hline
\tt functions-VotedPerceptron   & 0.333 & 0.015 & 0.978 & 0.333 & 0.473 & 0.102 & 0.783 \\ \hline
\tt bayes-NaiveBayes            & 0.760 & 0.250 & 0.971 & 0.760 & 0.844 & 0.175 & 0.772 \\ \hline
\tt Meta-RandomCommittee        & 0.572 & 0.193 & 0.971 & 0.572 & 0.706 & 0.114 & 0.771 \\ \hline
\hline
\end{tabular}
\end{center}
\label{default}
    \caption{\label{tab:featureClassifiers}
        We trained over 30 classifiers from \href{http://www.cs.waikato.ac.nz/ml/weka/}{Weka 3.8.0}
        on the features of MBBs listed in the main text to assess their performance.
        Top 10 classifiers in terms of \textit{ROC area} are reported below.
    }
\end{table*}

To close the discussion, let us assume that the system does indeed operate at the
EER operating point, conservatively, {\em EER=22\%}. Based on a random sampling
of the area, we estimate that the genuine features occupy $\frac{5}{64}$ portion
of the total area. If we assume that house artifact is pessimistically 
representative in terms of analytics performance and prevalence the experts will
be at least roughly $13$ times more productive while missing about $20\%$ of the
genuine artifacts.

\bibliographystyle{IEEEtran}
\bibliography{RSEPaper}

\end{document}